\documentclass[10pt,twocolumn,letterpaper]{article}

\usepackage{cvpr}              

\usepackage{graphicx}
\usepackage{amsmath}
\usepackage{amssymb}
\usepackage{booktabs}
\usepackage{multirow}
\usepackage[accsupp]{axessibility}  

\usepackage[colorlinks]{hyperref}

\usepackage[capitalize]{cleveref}
\crefname{section}{Sec.}{Secs.}
\Crefname{section}{Section}{Sections}
\Crefname{table}{Table}{Tables}
\crefname{table}{Tab.}{Tabs.}

\newcommand*\samethanks[1][\value{footnote}]{\footnotemark[#1]}
\def\cvprPaperID{4593} 
\def\confName{CVPR}
\def\confYear{2022}

\begin{document}

\title{Structural and Statistical Texture Knowledge Distillation for \\ Semantic Segmentation}

\author{
Deyi Ji\textsuperscript{\rm1,\rm2}\thanks{Equal Contribution}, 
~ Haoran Wang\textsuperscript{\rm1}\samethanks,  
~ Mingyuan Tao\textsuperscript{\rm1},  
~ Jianqiang Huang\textsuperscript{\rm1},  
~ Xian-Sheng Hua\textsuperscript{\rm1}\thanks{Corresponding Authors. Hongtao Lu is also with MOE Key Lab of Artificial Intelligence, AI Institute, Shanghai Jiao Tong University, China.}, 
~ Hongtao Lu\textsuperscript{\rm2}\samethanks\\
\textsuperscript{\rm1}Alibaba Cloud Computing Ltd. \\ 
\textsuperscript{\rm2}Department of Computer Science and Engineering, Shanghai Jiao Tong University \\
{\tt\small \{jideyi.jdy, jinglai.whr, juchen.tmy, jianqiang.hjq, xiansheng.hxs\}@alibaba-inc.com} \\ 
{\tt\small htlu@sjtu.edu.cn}
}
\maketitle

\begin{abstract}
Existing knowledge distillation works for semantic segmentation mainly focus on transferring high-level contextual knowledge from teacher to student. However, low-level texture knowledge is also of vital importance for characterizing the local structural pattern and global statistical property, such as boundary, smoothness, regularity and color contrast, which may not be well addressed by high-level deep features. In this paper, we are intended to take full advantage of both structural and statistical texture knowledge and propose a novel Structural and Statistical Texture Knowledge Distillation (SSTKD) framework for semantic segmentation. Specifically, for structural texture knowledge, we introduce a Contourlet Decomposition Module (CDM) that decomposes low-level features with iterative Laplacian pyramid and directional filter bank to mine the structural texture knowledge. For statistical knowledge, we propose a Denoised Texture Intensity Equalization Module (DTIEM) to adaptively extract and enhance statistical texture knowledge through heuristics iterative quantization and denoised operation. Finally, each knowledge learning is supervised by an individual loss function, forcing the student network to mimic the teacher better from a broader perspective. Experiments show that the proposed method achieves state-of-the-art performance on Cityscapes, Pascal VOC 2012 and ADE20K datasets.
\end{abstract}

\begin{figure}[!ht]
  \centering
  \includegraphics[width=0.97\linewidth]{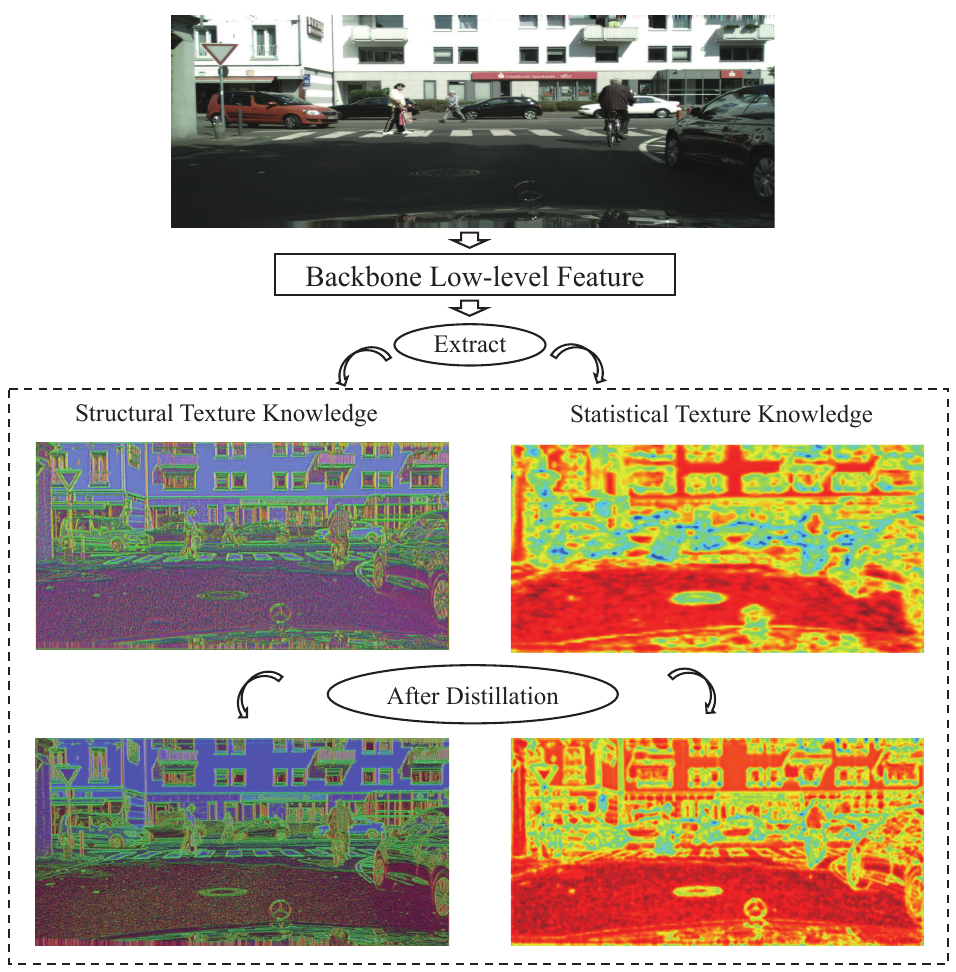}
  \caption{The overview of the structural and statistical texture knowledge distillation of an example image. Two kinds of the texture knowledge are extracted from the low-level feature of a CNN backbone. The original structural and statistical texture are fuzzy and in low-contrast. After distillation, the contour is clearer and the intensity contrast is more equalized, showing that two kinds of the texture are both enhanced.
  }
  \label{intro}
\end{figure}

\section{Introduction}
Semantic segmentation, which aims to assign each pixel a unique category label for the input image, is a crucial and challenging task in computer vision.
Recently, deep fully convolution network \cite{long2015fully} based methods have achieved remarkable results in semantic segmentation, and extensive methods have been investigated to improve the segmentation accuracy by introducing sophisticated models \cite{chen2018encoder,zhao2017pyramid,yuan2018ocnet,cdgcnet, ocrnet, stlnet}.
However, these methods are usually based on a large model, which contains tremendous parameters.
Since semantic segmentation has shown great potential in many applications like autonomous driving, video surveillance, robot sensing and so on, how to keep efficient inference speed and high accuracy with high-resolution images is a critical problem.

The focus of this paper is knowledge distillation, which is introduced by Hinton et al. \cite{hinton2015distilling} based on a teacher-student framework, and has received increasing attention in semantic segmentation community \cite{xie2018improving,he2019knowledge, liu2019structured, structure_dense, channel_dist}. Previous works mainly focus on the high-level contextual knowledge\cite{liu2019structured, channel_dist} or the final response knowledge\cite{xie2018improving,he2019knowledge, structure_dense}, which are appropriate to capture the global context and long range relation dependencies among pixels, but will also result in coarse and inaccurate segmentation results, since they are usually extracted  with a large receptive field and miss many low-level texture details. In this paper, we concentrate on exploiting the texture knowledge from the teacher to enrich the low-level information of the student. According to the digital image processing\cite{dip}, texture is a region descriptor which can provide measures for both local structural property and global statistical property of a image. The structural property can also be  viewed as spectral domain analysis and often refer to some local patterns, such as boundary, smoothness and coarseness. While the statistical property pay more attention to the global distribution analysis, such as histogram of intensity.

Based on the above analysis, we propose a novel Structural and Statistical Texture Knowledge Distillation (SSTKD) framework to effectively distillate two kinds of the texture knowledge from the teacher model to the student model, as shown in Figure \ref{intro}. More comprehensively,  we introduce a Contourlet Decomposition Module (CDM) that decomposes low-level features to mine the structural texture knowledge with iterative laplacian pyramid and directional filter bank. The contourlet decomposition is a kind of multiscale geometric analysis tool and can enable the neural network the ability of geometric transformations, thus is naturally suitable for describing the structural properties. Moreover, we propose a Denoised Texture Intensity Equalization module (DTIEM) to adaptively extract and enhance the statistical knowledge, cooperated with an Anchor-Based Adaptive Importance Sampler. The DTIEM can effectively describe the statistical texture intensity in a statistical manner in deep neural networks, as well as suppress the noise produced by the amplification effect in near-constant regions during the texture equalization.
Overall, our contribution is threefold:

\begin{itemize}
\setlength{\parskip}{0pt} \setlength{\itemsep}{0pt plus 1pt}
    \item To our knowledge, it is the first work to introduce both the structural and statistical texture to knowledge distillation for semantic segmentation. We propose  a novel Structural and Statistical Texture Knowledge Distillation (SSTKD) framework to effectively extract and enhance the unified texture knowledge and apply them to teacher-student distillation.
    \item More comprehensively, we introduce the Contourlet Decomposition Module (CDM) and propose the Denoised Texture Intensity Equalization Module (DTIEM) to describe the structural and statistical texture, respectively. Moreover, DTIEM utilizes an adaptive importance sampler and a denoised operation for efficient and accurate characterization.
    \item Experimental results show that the proposed framework achieves the state-of-the-art performance on three popular benchmark datasets in spite of the choice of student backbones.
\end{itemize}

\begin{figure*}[!ht]
    \centering
    \includegraphics[width=0.95\linewidth]{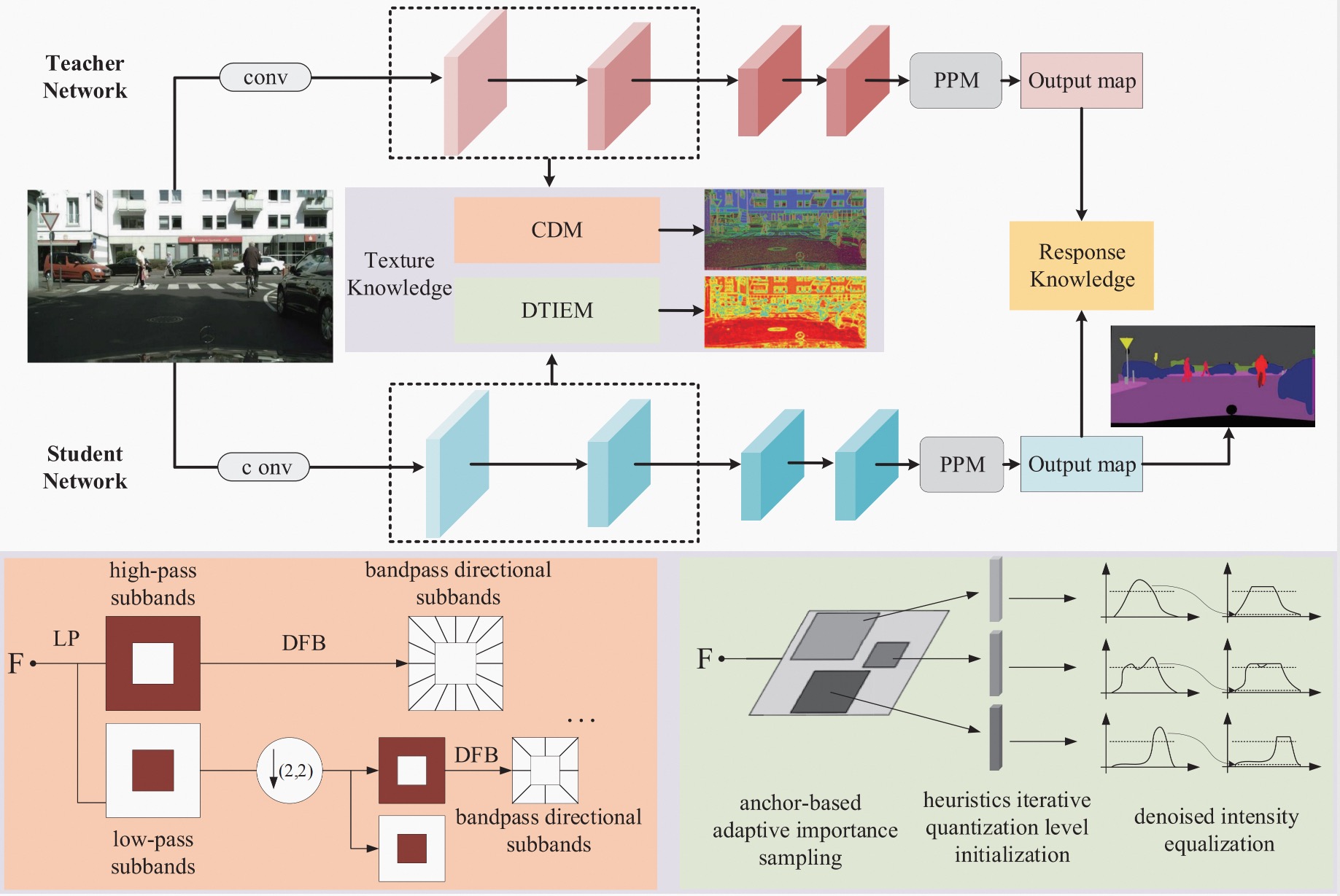}
    \caption{An overview of our proposed framework. PSPNet \cite{zhao2017pyramid} is used as the model architecture for both teacher and student network, which consists of the backbone network, pyramid pooling module (PPM) and the final output map. 
    Apart from the response knowledge, we further propose to extract the texture knowledge from low-level features. The corresponding parts of two kinds of the texture knowledge are depicted in the light red\cite{bamberger1992filter, c-cnn, andrearczyk2016using} and light green below the network pipeline, respectively.}
    \label{fig1}
\end{figure*}

\section{Related Work}
\noindent \textbf{Semantic Segmentation.}
In the past few years, CNN-based methods have achieved remarkable success in computer vision community \cite{szegedy2016inception,simonyan2014very,chen2018encoder,wang2021learning,ji2019end, feng2018challenges}. Existing works for semantic segmentation mainly follow 
fully convolutional networks (FCN) and focuse on contextual information and elaborate network. Many works \cite{zhao2017pyramid, chen2017deeplab, chen2016attention, lin2017exploring, cdgcnet, Ji_2023_CVPR} try to take advantage of rich contextual information in deep features, while the improvement of network design has also significantly driven the performance, such as graph modules \cite{velivckovic2017graph,cdgcnet, ji2020context} and attention techniques \cite{chen2016attention,fu2019dual,gpwformer,zagoruyko2016paying,ipgn}.
How to find a better trade-off between accuracy and efficiency have been discussed for a long time.
Real-time semantic segmentation algorithms aim to produce high-quality prediction under limited calculation \cite{zhao2018icnet, yu2018bisenet}.

\noindent \textbf{Knowledge Distillation.}
In a typical knowledge distillation framework, the logits are used as the knowledge from the teacher model \cite{hinton2015distilling}. Many works follow this paradigm in other visual applications, such as object detection and human pose estimation.
In semantic segmentation, existing methods also utilize this idea to produce fundamental results. They transfer the output maps to the student model, distilling the class probabilities for each pixel separately. Based on this, they further propose to extract the different knowledge from the task-specific view. He et al. \cite{he2019knowledge} propose an affinity distill module to transfer the long-rage dependencies among widely separated spatial regions from a teacher model to a student model. SKD \cite{liu2019structured} proposes structured knowledge distillation to transfer pairwise relations and holistic knowledge with the help of adversarial learning. IFVD \cite{wangintra2020} forces the student model to mimic the intra-class feature variation of the teacher model. CWD \cite{channel_dist} minimizes the Kullback–Leibler divergence between the channel-wise probability map of the teacher and student networks. Different from them, we first introduce the texture knowledge to semantic segmentation, showing an effective framework for this task.

\noindent \textbf{Texture in Semantic Segmentation.} In the perspective of digital image processing \cite{dip}, texture is a kind of descriptor that provides measures of properties such as smoothness, coarseness, regularity and so on. Image texture is not only about the local structural patterns, but also global statistical property. Zhu et al. \cite{stlnet} firstly introduce the statistical texture  to semantic segmentation, and propose a Quantizaton and Count Operator (QCO) to extract the low-level statistical texture feature, which is then aggregated with high-level contextual feature to obtain a more precise segmentation map. However the QCO performs as a global manner and sparse quantization, as well as neglects the noise produced by the amplification effect. Moreover, structural texture information is not well addressed in their work. In this paper, we focus on the unified texture information including both structural and statistical, and propose to improve the statistical characterization with anchor-based importance sampling, heuristics iterative quantization and denoised operation, as well as propose the CDM to re-emphasize the structural texture.

\section{Method}
In this section, we introduce the proposed Structural and Statistical Texture Knowledge Distillation (SSTKD) framework in detail. Firstly, we introduce the overall Structure in section \ref{overview}. Subsequently, we introduce the Structural Texture Knowledge Distillation and Statistical Texture Distillation in section \ref{Structural} and \ref{Statistical}, respectively. Finally, we provide the optimization process in section \ref{Optimization}.

\subsection{Overview} \label{overview}
The overall framework of the proposed method is illustrated in Figure \ref{fig1}. The upper network is the teacher network  while the lower one is the student network. Following the previous works\cite{liu2019structured, wangintra2020, channel_dist}, the PSPNet \cite{zhao2017pyramid} architecture is used for both the teacher and student, and ResNet-101 and ResNet-18 \cite{he2016deep} are used as their backbone respectively, which can be also changed to any other backbone networks. Firstly, we adopt the same basic idea in knowledge distillation to align the response-based knowledge between the teacher and student as previous works\cite{liu2019structured, wangintra2020, channel_dist}. Specifically, we use the KL divergence to supervise the pixel-wise probability distribution and adversarial learning to strengthen the output segmentation maps.
Furthermore, we propose to extract two kinds of the texture knowledge from the first two layers of ResNet backbone for both the teacher and student model, as the  texture information is more reflected on low-level features. For structural texture knowledge, we introduce a Contourlet Decomposition Module (CDM) which exploits structural information in the spectral space. For statistical texture knowledge, we introduce a Denoised Texture Intensity Equalization Module (DTIEM) to adaptively extract the statistical texture intensity histogram with an adaptive importance sampler, then enhance it with a denoised operation and graph reasoning. Finally we optimize two kinds of the knowledge between the teacher and student model with two individual Mean Squared (L2) loss.

\begin{figure}[!t]
    \centering
    \includegraphics[width=0.95\linewidth]{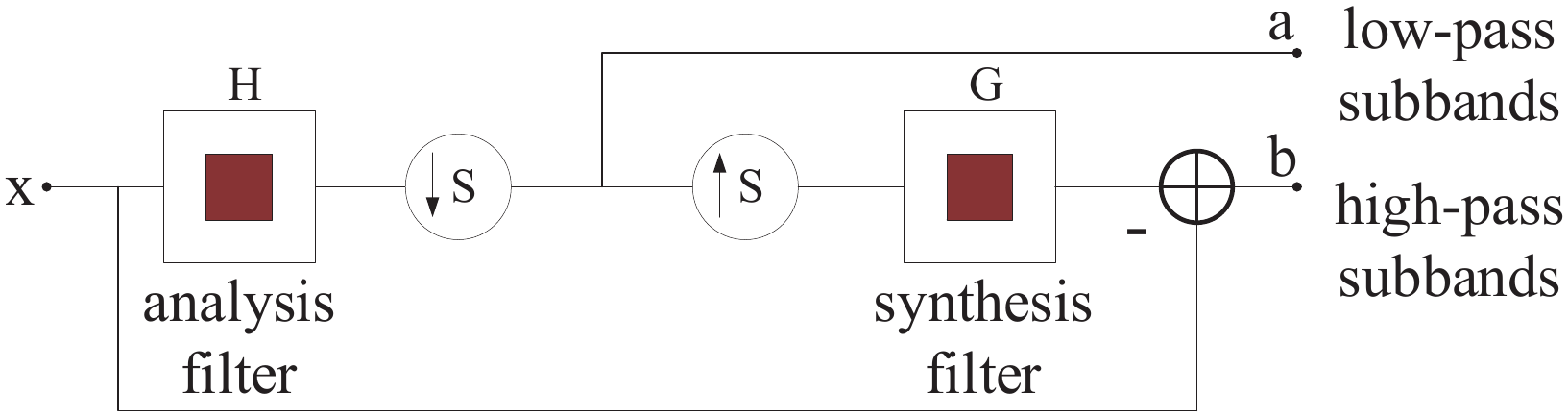}
    \caption{LP decomposition\cite{do2003contourlets, do2002contourlets, c-cnn}. The low-pass subbands $a$ is generated from the input $x$ with a low-pass analysis filters $H$ and a sampling matrix $S$. The high-pass subbands $b$ are then computed as the difference between $x$ and the prediction of $a$, with a sampling matrix $S$ followed by a low-pass synthesis filters $G$.}
    \label{lp}
\end{figure}

\subsection{Structural Texture Knowledge Distillation}\label{Structural}
Traditional filters have inherent advantages for texture representation with different scales and directions in the spectral domain, and here we consider to utilizing the contourlet decomposition, a kind of multiscale geometric analysis tool which has substantial advantages in locality and directionality\cite{do2005contourlet, donoho2001can, c-cnn, do2003contourlets, do2002contourlets, sha2005unsupervised}, and can enhance the ability of geometric transformations in CNN. Based on the advantages, we introduce a Contourlet Decomposition Module (CDM)\cite{do2002contourlets, c-cnn} to mine the texture knowledge in the spectral space.
The light red part in Figure \ref{fig1} shows the details of CDM. Specifically, it adopts a Laplacian Pyramid (LP) \cite{burt1983laplacian} and a directional filter bank (DFB) \cite{bamberger1992filter, andrearczyk2016using, cimpoi2015deep} iteratively on the low-pass image.
The LP is aimed to obtain multiscale decomposition. As shown in Figure \ref{lp}, given the input feature $x$, a low-pass analysis filters $H$ and a sampling matrix $S$ are used to generate downsampled low-pass subbands, then the high-pass subbands are obtained by the difference between the original $x$ and the intermediate result, which is computed by a sampling matrix $S$ and a low-pass synthesis filters $G$ \cite{burt1983laplacian, c-cnn}.
Next, DFB is utilized to reconstruct the original signal with a minimum sample representation, which is generated by $m$-level binary tree decomposition in the two dimensional frequency domain, resulting in $2^m$ directional subbands \cite{bamberger1992filter}. For example, the frequency domain is divided into $2^3=8$ directional subbands when $m=3$, and the subbands $0$-$3$ and $4$-$7$ correspond to the vertical and horizontal details, respectively.
Finally, the output of the contourlet decomposition in level $n$ can be described by the following equations:
\begin{equation}
    \begin{aligned}
        F_{l,n+1},F_{h,n+1}&=LP(F_{l,n})\downarrow p \\
        F_{bds,n+1}&=DFB(F_{h,n+1}) ~~~ n \in [1, m]
    \end{aligned}
\end{equation}
where the symbol $\downarrow$ is downsampling operator, $p$ denotes the interlaced downsampling factor, $l$ and $h$ represent the low-pass and high-pass components respectively, $bds$ denotes the bandpass directional subbands.

For a richer expression, we stack multiple contourlet decomposition layers iteratively in the CDM.
In this way, abundant bandpass directional features are obtained via the contourlet decomposition from the low-level features, which are used as the structural texture knowledge $F^{str}$ for distillation. 
We apply the CDM to the teacher and student networks respectively and use the conventional Mean Squared (L2) loss to formulate the texture distillation loss:
\begin{equation}
        L_{str}(S)=\frac{1}{(W \times H)}\sum_{i\in R}{(F^{str;T}_{i}-F^{str;S}_{i})^2}
\end{equation}
where $F^{str;T}_{i}$ and $F^{str;S}_{i}$ denote the $i$th pixel in texture features produced from the teacher network $T$ and the student network $S$, and $i \in R=W \times H$ represents the feature size.

\subsection{Statistical Texture Knowledge Distillation}\label{Statistical}

The statistical texture is usually of a wide variety and a continuous distribution in spectral domain, which is difficult to be extracted and optimized in deep neural networks. Previous work \cite{stlnet} firstly proposed a Quantization and Count Operator (QCO) to describe the statistical texture, which quantized the whole input feature into multiple uniform levels, then counted the number of features belonging to each level to get the quantization encoding matrix, followed by a graph module to perform quantization levels redistribution and texture enhancement. However, there are three limitations in the ordinary QCO. Firstly, it quantizes the input feature in a global manner, resulting in very sparse and discrete quantizaton levels distribution, which can not well balance the trade-off between accurate texture and computation burden. Secondly, amounts of pixels will never be quantized to any level as it generates the initial quantization levels by a uniform distribution, meanwhile using several narrow-peak functions to quantize the input feature. Thirdly, it may come across quantization noise produced by the over-amplify effect in near-constant areas during quantization process.

We propose to describe the statistical knowledge based on the ordinary QCO. Moreover, we propose three improvements in response to the three limitations as shown in the light green part in Figure \ref{fig1}. Firstly, for the sake of accurate characterization and efficient computation, we exploit an Anchor-Based Adaptive Importance Sampler to only select the important regions for extraction. Secondly, we design a heuristics iterative method for a more equalized quantization level initialization. Thirdly, we utilize a denoised operation to suppress the over-amplify effect. Based on above analysis, we propose the Denoised Texture Intensity Equalization Module (DTIEM) to extract and enhance the statistical texture adaptively. In the following sections, we illustrate the above modules in detail.

\noindent\textbf{Anchor-Based Adaptive Importance Sampler.} \label{anchor_sampler}
Extracting the statistical texture of the whole input feature is straightforward, but it lacks the attention on discriminative regions and only works well when the feature intensities of pixels are near-uniformly distributed. However, real-world scenes are usually offended with chaotic conditions and the pixel intensities tend to be heavily imbalanced distributed. Moreover, the global operation on a whole image is not able to describe local intensity contrast accurately and always means a large computation burden. Thus we consider using an importance sampling method to mine the hard-to-classify areas, where the feature intensity distribution is discrete and varying, while the statistical texture is rich and diverse. Following the typical paradiam\cite{pointrend}, we design an Anchor-Based Adaptive Importance Sampler. It is designed to bias selection towards most uncertain regions, while retaining some degree of uniform coverage, by the following steps \cite{pointrend}. (i) \textit{Over Generation}: Aiming at sample $M$ points, to guarantee the variety and recall, we over-generate candidate points by randomly selecting the $kM(k>1)$ points with a uniform distribution. (ii) \textit{Importance sampling:} In the $kM$ points, we are intended to choose out the most uncertain $\beta M(\beta \in [0, 1])$ ones with an anchor-based adaptive importance sampling strategy. (iii) \textit{Coverage}: to balance the distribution, we select the remaining $(1-\beta) M$ ones from the rest of points with a uniform distribution.

As found that the most essential step lies in  \textit{importance sampling}, we formulate this process as follows. For each sample $s_i \in kM$, we set several anchors with various scales and aspect ratios at the location. In this way, we generate $\xi$ region proposals $R_i$ for each $s_i$. 
For each $r_{ij} \in R_i (i\in [1, kM] , j\in [1, \xi])$, we calculate its sample probability by:
\begin{equation}
\begin{aligned}
    prob_{ij} = std(r_{ij})
\end{aligned}
\end{equation}
\noindent where $std(\cdot)$ means the variance function. It shows that the region with larger variance will be more likely sampled, as its intensity distribution is diverse and may have more rich statistical textures that need to be enhanced. Noted that the student utilizes the same importance sampling results as teacher to select region proposals.

\noindent\textbf{Texture Extraction and Intensity Equalization.}

QCO is inspired of histogram quantification\cite{dip} and describes texture in a statistical manner. Given the input feature $\mathbf{A} \in \mathbb{R}^{C\times H\times W}$, it firstly calculates the self-similarity matrix $\textbf{S}$, and quantizes it into $N$ uniform levels $\textbf{L}$. Next it counts the value for each quantization level and gets the quantization encoding matrix $\textbf{E} \in \mathbb{R}^{N\times HW}$. Finally it utilizes a simple fully-connected graph to perform quantization levels redistribution and texture enhancement. The quantization can be formulated as:

\begin{equation}
\begin{aligned}
    &\textbf{L}_{n} = \frac{max\left(\mathbf{S}\right) - min\left(\mathbf{S}\right)}{N}\cdot n + min(\mathbf{S}), \ \   n \in [1, N]\\
    &\mathbf{E}^{n}_{i} = \left\{
        \begin{array}{lcc}
        1-|L_{n}-\mathbf{S}_{i}|,  & if & -\frac{0.5}{N}\le L_{n}-\mathbf{S}_{i}<\frac{0.5}{N}\\
        0 & & else
    \end{array}
    \right.
    \label{qfunc}
    \end{aligned}
\end{equation}

\noindent where $i\in[1, HW]$. 
Based on the above discussion, we propose the other two improvements here.

For the quantization level, a heuristics iterative method is adopted instead of a uniform quantization. Specifically, we use a simple $\textit{t}$-step uniform sampling to formulate an attention sampled results. Firstly, we over-quantize the input into $2N$ levels to guarantee that most points can be quantized to one of the levels, obtaining the quantization encoding matrix with Eq. \ref{qfunc}. Based on the count values of all quantization levels, we set an intensity ratio threshold $\delta$ to divide the quantization levels into two groups $G_{<\delta}$ and $G_{>=\delta}$, which are then re-quantized into $\alpha N(\alpha \in [0,1])$ and $(1-\alpha) N$ levels to obtain final $N$ levels. The above process can be performed iteratively. In this way, the imbalanced distribution problem among quantization levels can be weaken. Moreover, for each quantization function in Eq. \ref{qfunc}, we widen its peak adaptively by replacing the $N$ with its current group levels number, as each intermediate group level number is much smaller than $N$, thus the cover range of each quantization function can be broadened.

\noindent\textbf{Intensity-Limited Denoised Strategy.} For the noise overamplity problem, inspired by the Contrast-Limited Adaptive Histogram Equalization \cite{ahe}, we propose an intensity-limited denoised strategy to constrain the intensity peak and redistribute the extra peak to all quantization levels dynamically. More comprehensively, for each selected region and $N$ quantization levels, we get the initial quantization encoding matrix $\textbf{E}$ by Eq. \ref{qfunc}. Due to the near-constant sub-area, the count value of some levels may show in a very high peak \cite{ahe}, we perform an intensity-clip operation to limit these peaks with a given ratio threshold $\theta$, and then redistribute the sum of extra peak values $\textbf{E}_{extra}$ equally among all the quantization levels. Finally, the denoised quantization encoding matrix $\textbf{DE}$ can be calculated:

\begin{equation}
    \textbf{E}_{extra} = \sum_{n}^{N} [max(\textbf{E}^n - \theta \cdot max(\textbf{E}), 0)]
\end{equation}

\begin{equation}
    \mathbf{DE}^{n} = \left\{
        \begin{array}{lcc}
        \theta \cdot max(\textbf{E}) + \frac{\textbf{E}_{extra}}{N} &if& \textbf{E}^{n} > \theta \cdot max(\textbf{E}) \\
        \textbf{E}^{n} + \frac{\textbf{E}_{extra}}{N} & & else
    \end{array}
    \right.
    \label{Edfunc}
\end{equation}
\noindent where $n \in [1, N]$, $max(\cdot)$ is the maximum function. Then $\textbf{DE}$ is also enhanced with the graph reasoning as the previous work\cite{stlnet} to get the statistical texture knowledge $F^{sta}$. 
Similarly as the structural texture knowledge distillation, we apply the DTIEM to both teacher and student respectively and use the L2 loss to formulate the distillation loss:
\begin{equation}
        L_{sta}(S)=\frac{1}{(W \times H)}\sum_{i\in R}{(F^{sta;T}_{i}-F^{sta;S}_{i})^2}
\end{equation}
\noindent where $F^{sta;T}_{i}$ and $F^{sta;S}_{i}$ denote the statistical texture knowledge of the teacher and student respectively.

\subsection{Optimization} \label{Optimization}

Following the common practice and previous knowledge distillation works for semantic segmentation \cite{liu2019structured, wangintra2020}, we also add the fundamental response-based distillation loss $L_{re}$ and adversarial loss $L_{adv}$ for stable gradient descent optimization:
\begin{equation}
    L_{re} = \frac{1}{(W_{re} \times H_{re})} \sum_{i\in R} KL(P_i^{re;T} || P_i^{re;S})
\end{equation}
where $P_i^{re;T}$ and $P_i^{re;S}$ denote the class probabilities of $i$-th pixel produced by the teacher and student model respectively, and $i \in R=W_{re} \times H_{re}$ represents the output size. The adversarial training is aimed to formulate the holistic distillation problem  \cite{xu2017training, liu2019structured} and  denoted as follows:
\begin{equation}
    L_{adv} = \mathbb{E}_{S \sim p(S)}[D(S|I)]
\end{equation}
where $\mathbb{E}(\cdot)$ is the expectation operator, and $D(\cdot)$ is the discriminator. $I$ and $S$ are the input image and corresponding segmentation map respectively.

Therefore,  for the overall optimization, the whole objective function consists of a conventional cross-entropy loss $L_{seg}$  for semantic segmentation and the above-mentioned distillation loss:
\begin{equation}
\begin{aligned}
    L = L_{seg} + \lambda_1L_{str} + \lambda_2L_{sta} + \lambda_3L_{re} - \lambda_4 L_{adv}
\end{aligned}
\end{equation}
where $\lambda_1, \lambda_2, \lambda_3, \lambda_4$ are set to 0.9, 1.15, 5, 0.01, respectively.

\section{Experiments}
\subsection{Datasets and Evaluation Metrics}
To verify the effectiveness of the proposed method, we conduct experiments on the following large-scale datasets.
\noindent \textbf{Cityscapes.} The Cityscapes dataset\cite{cordts2016cityscapes} has 5,000 images captured from 50 different cities, and contains 19 semantic classes.
Each image has $2048\times1024$ pixels, which have high quality pixel-level labels of 19 semantic classes.
There are 2,979/500/1,525 images for training, validation and testing.

\noindent \textbf{ADE20K.} The ADE20K dataset has 20K/2K/3K images for training, validation, and testing, and contains 150 classes of diverse scenes.

\noindent \textbf{Pascal VOC 2012.} The Pascal VOC 2012 dataset\cite{everingham2010pascal} is a segmentation benchmark of 10,582/1,449/1,456 images for training, validation and  testing, which involves 20 foreground object classes and one background class.

\noindent \textbf{Evaluation Metrics.} In all experiments, we adopt the mean Intersection-over-Union (mIoU) to study the distillation effectiveness.
The model size is represented by the number of network parameters, and the Complexity is evaluated by the sum of floating point operations (FLOPs) in one forward propagation on a fixed input size.

\subsection{Implementation Details}
Following \cite{liu2019structured,wangintra2020,channel_dist}, we adopt PSPNet \cite{zhao2017pyramid} with ResNet101 \cite{he2016deep} backbone as the teacher network, and use PSPNet with different compact backbones as the student networks, including Resnet18\cite{he2016deep} and EfficientNet-B1 \cite{tan2019efficientnet}, which also validates the effectiveness when the teacher model and the student model are of different architectural types. In this paper, we adopt an offline distillation method, first train the teacher model and then keep the parameters frozen during the distillation progress.
In the training process of the student network, random scaling (from 0.5 to 2.1) and random horizontal flipping (with the probability of 0.5) are applied as the data augmentation. We implement two-level contourlet decomposition iteratively in the CDM, where $m$ is set to $4$ and $3$ respectively. We set $N=50, \alpha=0.3, \theta=0.9$. Stochastic Gradient Descent with momentum is deployed as the optimizer, where the momentum is 0.9 and weight decay rate is 1e-5. The base learning rate is 0.015 and multiplied by $(1-\frac{iter}{max-iter})^{0.9}$.  We train the model for 80000 iterations with the batch size of 16.

\subsection{Ablation Study}

In all the ablation studies, we use Cityscapes validation dataset, and ResNet-18 pretrained from ImageNet as the backbone of the student network.

\begin{table}[!t]
\centering
\scalebox{1}{\begin{tabular}{l|c}
\hline
Method & mIOU(\%) \\ \hline
T: PSPNet-R101 &  78.56 \\
S: PSPNet-R18 & 69.10 \\ \hline
+Response Knowledge & 72.47 \\
+Response+Structural Texture Knowledge & 74.10  \\
+Response+Statistical Texture Knowledge & 74.69 \\
& \\
\multirow{-2}{*}{\begin{tabular}[l]{@{}l@{}}+Response+Structural+Statistical\\ \ \ Texture Knowledge\end{tabular}} & \multirow{-2}{*}{75.15} \\ \hline
\end{tabular}}
\caption{Efficacy of two kinds of the texture knowledge.}
\label{efficacy_of_texture}
\end{table}

\begin{table}[!t]
\centering
 \scalebox{1}{
    \begin{tabular}{c|c}
    \hline
    Levels number in CDM  &\textit{val} mIoU(\%) \\ \hline
    baseline & 72.47 \\ \hline
    1 & 73.44 \\
    2 & 74.10 \\
    3 & 74.11 \\
    \hline
    \end{tabular}
    }
\caption{The impact of layer number of comtourlet decomposition.}
\label{layer_num}
\end{table}

\begin{table}[!t]
\centering
 \centering
    \scalebox{1}{
        \begin{tabular}{c|c|c}
        \hline
        Method  & Params (M)  & FLOPs (G)  \\ \hline
        PSPNet & 70.43 & 574.9 \\ \hline
        CDM & 1.24 & 10.90 \\
        DTIEM & 2.80 & 23.73 \\
        \hline
    \end{tabular}
    }
    \caption{The FlOPs and Parameters of the proposed texture modules.}
    \label{flop}
\end{table}

\begin{table}[!t]
\centering
\scalebox{0.95}{\begin{tabular}{l|c}
\hline
Components in DTIEM & mIOU(\%) \\ \hline
baseline & 72.47 \\ \hline
+Global &  73.57 \\
+Adap. Samp. &  74.06 \\
+Adap. Samp. + Heuristics Init. & 74.25 \\
+Adap. Samp. + Heuristics Init. + Denoised & 74.69 \\ \hline
\end{tabular}}
\caption{Abalation Study of Statistiacal Knowledge. ``Global" means global operation without sampling, ``Adap. Samp." means anchor-based adaptive importance sampling, ``Heuristics Init." means quantization levels heuristics initialization.}
\label{alation_ststistical}
\end{table}

\noindent \textbf{Efficacy of Two Kinds of the Texture Knowledge.}
Table \ref{efficacy_of_texture} shows the effectiveness of two kinds of the texture knowledge. The student network without distillation achieves the result of 69.10\%, and the response knowledge improves it to 72.47\%. Then we further add two kinds of the texture knowledge successively to validate the effect of each one. Concretely, it shows that the structural texture knowledge brings the improvement to 5.0\%, and the statistical texture knowledge  boosts the improvement to 5.59\%. Finally, when we add both the texture knowledge, the performance is promoted to 75.15\% with a large increase of 6.05\%. The gap between student and teacher is finally reduced, providing a closer result to the teacher network.

\noindent \textbf{Analysis of Structural Texture Knowledge.}
We conduct experiments to verify the effectiveness of the components in the structural texture module, and show the impact of level number of comtourlet decomposition in the CDM. As shown in Table \ref{layer_num}, ``baseline" means the results of student network with response knowledge,  with the level number gradually increasing,  mIOU is gradually increasing and stays still around $74.1\%$, which indicates that the texture knowledge is almost saturated when the level number gets to $2$.

\begin{figure}[!h]
    \centering
    \includegraphics[width=1\linewidth]{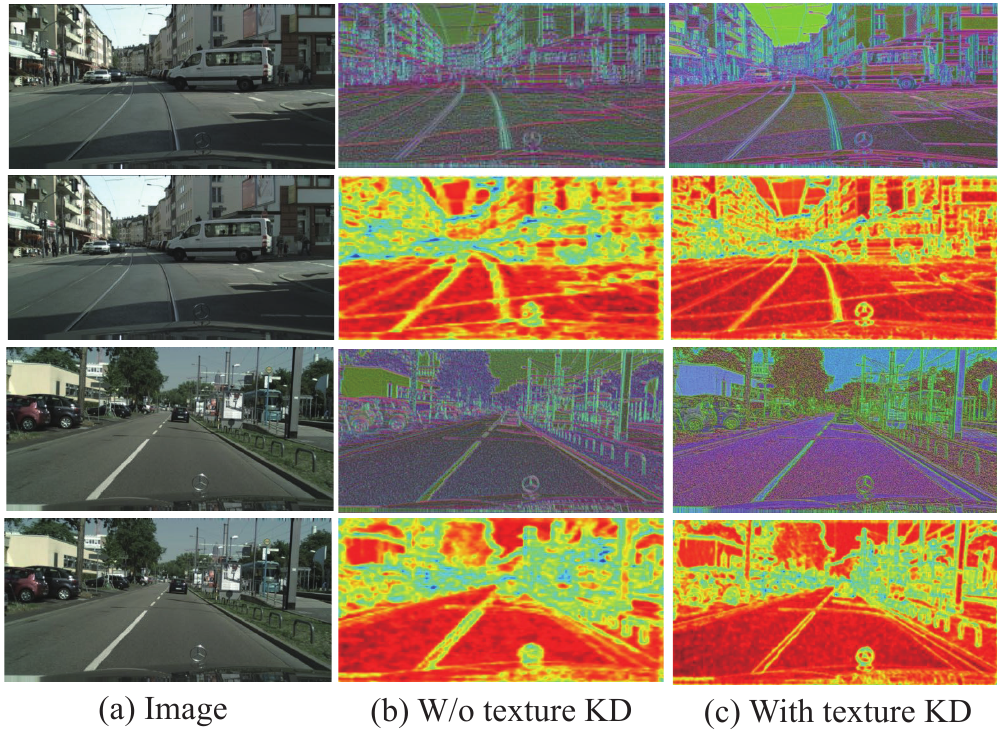}
    \caption{Comparison of visualization of the low-level feature from stage 1 of the backbone. KD means knowledge distillation. (a) is the original image. (b) is from the student network without texture knowledge distillation, (c) shows the changes after applying it in our method. Line 1 and 3 show the structural texture, while line 2 and 4 show the statistical texture.}
    \label{texture_vis}
\end{figure}

\noindent \textbf{Analysis of Statistical Texture Knowledge.}
In Table \ref{alation_ststistical}, we show the effectness of different components in the statistical texture module, ``baseline" means the results of student network with response knowledge. Firstly, we perform a global feature intensity equalization on the whole image and get a limited improvement. Then we add the adaptive sampling,   quantization level heuristics initialization and the denoised strategy successively, and the performance is gradually improved in this case.

\noindent \textbf{Complexity of Texture Extraction Modules.}
We show that the proposed texture modules are lightweight in Table \ref{flop}, which are estimated with the fixed input size. It shows that the CDM and DTIEM only bring very little extra cost compared to the PSPNet.

\begin{figure}[!t]
    \centering
    \includegraphics[width=1\linewidth]{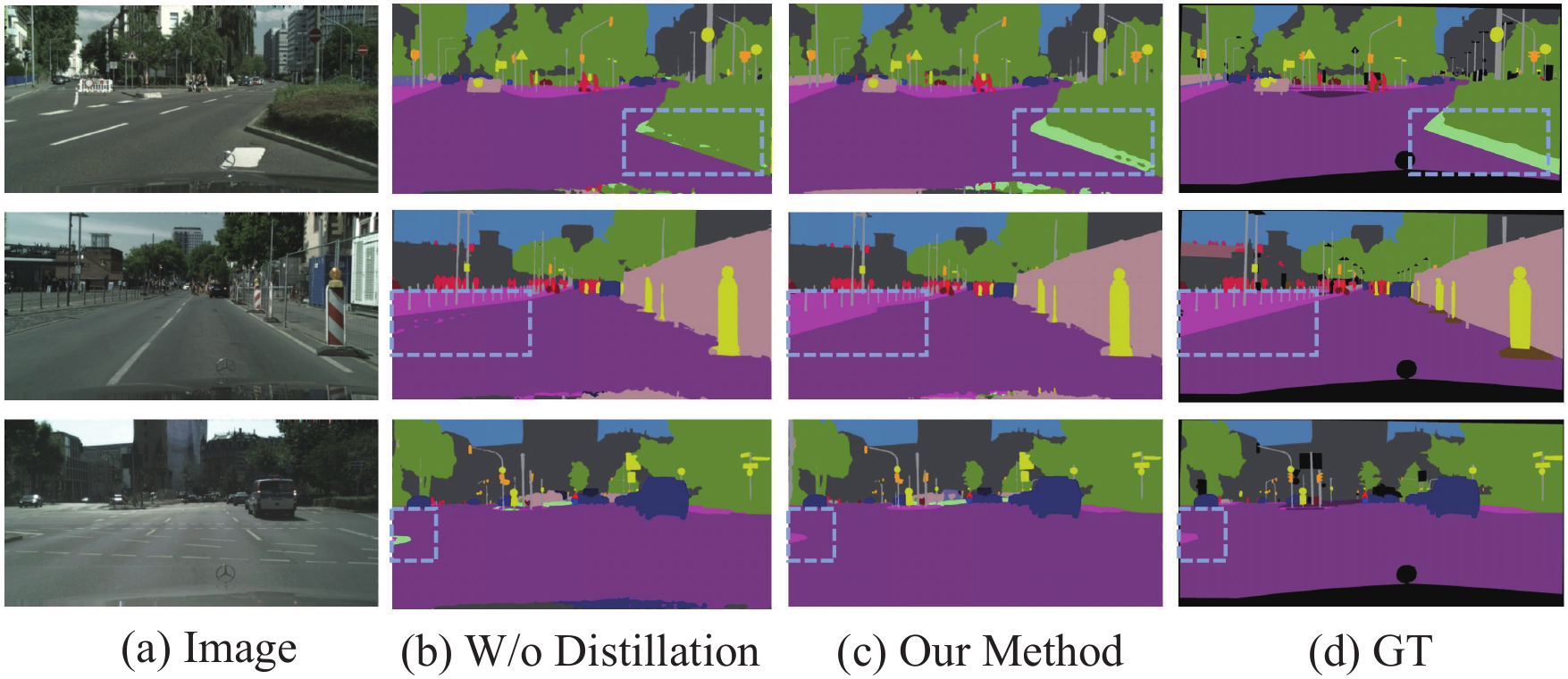} 
    \caption{Visual improvements on Cityscapes dataset: (a) orginal images, (b) w/o distillation, (c) Our distillation method, (d) ground truth. Our method improves the student network w/o distillation to produce more accurate and detailed results, which are circled by dotted lines.}
    \label{city_vis}
\end{figure}

\noindent \textbf{Visualization.}
Figure \ref{texture_vis} shows the visualization of the low-level feature from the student network with ResNet18 backbone. For comparison, we give the results of the student network w/ and w/o the texture knowledge distillation in our method. Specifically, Figure \ref{texture_vis} (b) shows the results which are produced without the structural/statistical texture knowledge. As we can see, the texture details are very fuzzy and in low-contrast, the contours of the objects are also incomplete. By contrast, they are clearer in Figure \ref{texture_vis} (c) when incorporating with two kinds of the texture knowledge. In this case, the contours of the auto logo and lane lines can offer more correct cues for semantic segmentation. The phenomenon illustrates the validity of the texture knowledge, providing a better understanding of our method. Besides, we also visualize the results of the different methods in Figure \ref{city_vis} for a better comparison.

\subsection{Comparison with State-of-the-arts}

\begin{table}
\centering
\scalebox{0.95}{\begin{tabular}{l|cc|c|c}
\hline
\multirow{2}{*}{Method} & \multicolumn{2}{c|}{\begin{tabular}[c]{@{}c@{}}Cityscapes\\ mIOU(\%)\end{tabular}} & \multirow{2}{*}{\begin{tabular}[c]{@{}c@{}}Params\\ (M)\end{tabular}} & \multirow{2}{*}{\begin{tabular}[c]{@{}c@{}}Flops\\ (G)\end{tabular}} \\ \cline{2-3}
                        & \multicolumn{1}{c|}{\textit{val}}  & \textit{test}  &        &         \\ \hline
ENet\cite{paszke2016enet}  & \multicolumn{1}{c|}{-}    & 58.3  & 0.358  & 3.612   \\ \hline
ICNet\cite{zhao2018icnet} & \multicolumn{1}{c|}{-}    & 69.5  & 26.50  & 28.30   \\ \hline
FCN\cite{long2015fully} & \multicolumn{1}{c|}{-}    & 62.7  & 134.5  & 333.9   \\ \hline
RefineNet\cite{lin2017refinenet}  & \multicolumn{1}{c|}{-}    & 73.6  & 118.1  & 525.7   \\ \hline
OCNet\cite{yuan2018ocnet}  & \multicolumn{1}{c|}{-}     & 80.1  & 62.58  & 548.5     \\ \hline
STLNet\cite{stlnet} & \multicolumn{1}{c|}{82.3} & 82.3  & 81.39  & 293.03  \\ \hline
\hline
\multicolumn{5}{c}{Results w/ and w/o distillation schemes} \\ \hline
\hline
T: PSPNet-R101\cite{zhao2017pyramid} & \multicolumn{1}{c|}{78.56} & 78.4 & 70.43& 574.9 \\ \hline
S: PSPNet-R18 & \multicolumn{1}{c|}{69.10} & 67.60 & 13.07 & 125.8 \\
+ SKDS \cite{liu2019structured}& \multicolumn{1}{c|}{72.70}& 71.40 & 13.07 & 125.8 \\
+ SKDD \cite{structure_dense}& \multicolumn{1}{c|}{74.08 }& - & 13.07 & 125.8 \\
+ IFVD \cite{wangintra2020} & \multicolumn{1}{c|}{74.54}& 72.74 & 13.07 & 125.8 \\
+ CWD \cite{channel_dist} & \multicolumn{1}{c|}{74.87}& - & 13.07 & 125.8 \\
+ SSTKD  & \multicolumn{1}{c|}{75.15} & 74.39 & 13.07 & 125.8 \\ \hline
S: Deeplab-R18 & \multicolumn{1}{c|}{73.37} & 72.39 & 12.62 & 123.9 \\
+ SKDS \cite{liu2019structured}& \multicolumn{1}{c|}{73.87}& 72.63 & 12.62 & 123.9 \\
+ IFVD \cite{wangintra2020}& \multicolumn{1}{c|}{74.09 }& 72.97 & 12.62 & 123.9 \\
+ CWD \cite{channel_dist} & \multicolumn{1}{c|}{75.91}& 74.32 & 12.62 & 123.9 \\
+ SSTKD  & \multicolumn{1}{c|}{76.13} & 75.01 & 12.62 & 123.9 \\ \hline
S: EfficientNet-B1 & \multicolumn{1}{c|}{60.40} & 59.91 & 6.70 & 9.896 \\
+ SKDS \cite{liu2019structured}& \multicolumn{1}{c|}{63.13}& 62.59 & 6.70 & 9.896 \\
+ IFVD \cite{wangintra2020}& \multicolumn{1}{c|}{66.50}& 64.42 & 6.70 & 9.896 \\
+ CWD \cite{channel_dist} & \multicolumn{1}{c|}{-}& - & 6.70 & 9.896 \\
+ SSTKD  & \multicolumn{1}{c|}{68.26} & 65.77 & 6.70 & 9.896 \\ \hline
\end{tabular}}
\caption{Quantitative results on Cityscapes. ``R18"(``R101") means ResNet-18(ResNet-101).} 
\label{sotacity}
\end{table}

\begin{table}
\centering
\scalebox{0.95}{\begin{tabular}{l|c|c|c}
\hline

Method & \begin{tabular}[c]{@{}c@{}}Pascal VOC \\ mIOU(\%)\end{tabular}  & \begin{tabular}[c]{@{}c@{}}ADE20K\\ mIOU(\%)\end{tabular} & \begin{tabular}[c]{@{}c@{}} Params \\ (M) \end{tabular} \\ \hline
FCN\cite{long2015fully}  &  69.6  & 39.91   & 134.5 \\ \hline
RefineNet\cite{lin2017refinenet}  &  82.4  & 40.7    & 118.1 \\ \hline
Deeplab V3\cite{chen2017deeplab}    &  77.9  & 44.99   & 87.1 \\ \hline
PSANet\cite{psanet}                 &  77.9  & 43.47   & 78.13 \\ \hline
OCRNet \cite{ocrnet}                 &  80.3  & 43.7    & 70.37 \\ \hline
\hline
\multicolumn{4}{c}{Results w/ and w/o distillation schemes} \\ \hline
\hline
T: PSPNet-R101\cite{zhao2017pyramid} & 78.52 & 44.39 & 70.43\\ \hline
S: PSPNet-R18& 65.42 & 24.65 & 13.07 \\
+SKDS \cite{liu2019structured} & 67.73 & 25.11 &  13.07 \\
+IFVD \cite{wangintra2020} & 68.04 & 25.72 &  13.07 \\
+CWD \cite{structure_dense} & 69.25 & 26.80 &  13.07 \\
+SSTKD & 70.98 & 29.19 &  13.07 \\ \hline
S: Deeplab-R18 & 66.81 & 24.89 & 12.62 \\
+SKDS \cite{liu2019structured} & 68.13 & 25.52 & 12.62 \\
+IFVD \cite{wangintra2020} & 68.42 & 26.53 &  12.62 \\
+CWD \cite{structure_dense} & 69.97 & 27.37 &  12.62 \\
+SSTKD & 71.45 & 29.79 & 12.62 \\ \hline
\end{tabular}}
\caption{Quantitative results on Pascal VOC 2012 and ADE20K. ``R18"(``R101") means ResNet-18(ResNet-101)}. 
\label{sota_voc_ade}
\end{table}

\noindent \textbf{Cityscapes.}
Table \ref{sotacity} shows that the proposed SSTKD framework achieves state-of-the-art results with different backbones in Cityscapes validation and test datasets. More comprehensively, SSTKD improves the student model(PSPNet) built on ResNet-18 to 75.15\% and 74.39\% on validation and test dataset respectively.
Moreover, we also change the student backbone to Deeplab and EfficientNet-B1, which shows the generality of SSTKD. Besides, the experimental results also show that we improve the baseline by a large margin.

\noindent \textbf{Pascal VOC 2012 and ADE20K.}
Table \ref{sota_voc_ade} shows the comparisons with state-of-the-art methods on PASCAL VOC 2012 and ADE20K validation dataset. Experimental results show that the proposed method improves the performance of the student network without distillation, meanwhile surpasses previous works in all cases in spite of the choice of architectures and backbones for student networks.

\section{Conclusion}
In this paper, we focus on the low-level structural and statistical knowledge in distillation for semantic segmentation. Specifically, we introduce the Contourlet Decomposition Module to effectively extract the structural texture knowledge, and the Denoised Texture Intensity Equalization Module to describe and enhance statistical texture knowledge, respectively. Under different supervisions, we force the student network to mimic the teacher network better from a broader perspective. Experimental results show that we achieve new state-of-the-art results on three semantic segmentation datasets, which proves the effectiveness and superiority of our method.

\section*{Acknowledgements}

This work was partially supported by the National Key R\&D Program of China under Grant 2020AAA0103902, NSFC(No. 62176155 and No. 61772330), Science and Technology Commission of Shanghai Municipality under Grant 2021SHZDZX0102.

{\small
\bibliographystyle{unsrt}
\bibliography{egbib}
}

\end{document}